\documentclass[journal]{IEEEtai}

\usepackage[colorlinks,urlcolor=blue,linkcolor=blue,citecolor=blue]{hyperref}

\usepackage{color,array}
\usepackage{amssymb}
\usepackage{amsmath}
\usepackage{txfonts}
\usepackage[classicReIm]{kpfonts}
\usepackage{mathdots}
\usepackage{graphicx}
\usepackage{algorithm}
\usepackage{algpseudocode}
\usepackage{indentfirst}
\usepackage{makecell}

\begin{document}

\title{Robust Convergence in Federated Learning through Label-wise Clustering}

\author{Hunmin Lee, Yueyang Liu, \IEEEmembership{Graduate Student Member, IEEE}, Donghyun Kim, \IEEEmembership{Senior, IEEE}, and Yingshu Li, \IEEEmembership{Senior, IEEE}
\thanks{This paragraph of the first footnote will contain the date on which you submitted your paper for review.}
\thanks{Hunmin Lee is with the Department of Computer Science, Georgia State University, Atlanta, GA 30302 USA (e-mail: hlee185@student.gsu.edu).}
\thanks{Yueyang Liu is with the Department of Computer Science, Georgia State University, Atlanta, GA 30302 USA (e-mail: yliu114@student.gsu.edu).}
\thanks{Donghyun Kim is with the Department of Computer Science, Georgia State University, Atlanta, GA 30302 USA (e-mail: dhkim@gsu.edu).}
\thanks{Yingshu Li is with the Department of Computer Science, Georgia State University, Atlanta, GA 30302 USA (e-mail: yili@gsu.edu).}
\thanks{This paragraph will include the Associate Editor who handled your paper.}
}

\markboth{Journal of IEEE Transactions on Artificial Intelligence, Vol. 00, No. 0, Month 2020}
{H. Lee \MakeLowercase{\textit{et al.}}: Robust Convergence in Federated Learning through Label-wise Clustering}

\maketitle

\begin{abstract}
Non-IID dataset and heterogeneous environment of the local clients are regarded as a major issue in Federated Learning (FL), causing a downturn in the convergence without achieving satisfactory performance. In this paper, we propose a novel Label-wise clustering algorithm that guarantees the trainability among geographically dispersed heterogeneous local clients, by selecting only the local models trained with a dataset that approximates into uniformly distributed class labels, which is likely to obtain faster minimization of the loss and increment the accuracy among the FL network. Through conducting experiments on the suggested six common non-IID scenarios, we empirically show that the vanilla FL aggregation model is incapable of gaining robust convergence generating biased pre-trained local models and drifting the local weights to mislead the trainability in the worst case. Moreover, we quantitatively estimate the expected performance of the local models before training, which offers a global server to select the optimal clients, saving additional computational costs. Ultimately, in order to gain resolution of the non-convergence in such non-IID situations, we design clustering algorithms based on local input class labels, accommodating the diversity and assorting clients that could lead the overall system to attain the swift convergence as global training continues. Our paper shows that proposed Label-wise clustering demonstrates prompt and robust convergence compared to other FL algorithms when local training datasets are non-IID or coexist with IID through multiple experiments.
\end{abstract}

\begin{IEEEImpStatement}
As contemporary systems are being operated in dynamic situations which are alternating into decentralized and distributed environments from conventional centralized frameworks, Federated Learning (FL) has been gaining attention for it is an optimal architecture when aggregating the information of geographically distributed dataset. However, the convergence of FL suffers to skewed and biased non-IID local dataset which are common in real-world practice. Based on the basic but powerful intuition of selecting the optimal clients that approximates to the uniform distribution of local class labels, our label-wise clustering algorithm provides robust convergence when learning among distributed local models in FL. Our paper suggests the six types of common non-IID scenarios in FL and limitation analysis when those cases are trained in vanilla FL. We also propose the evaluation methodology that quantitatively assess the local models through statistical properties without training them, which offers the significant preservation of resources.
\end{IEEEImpStatement}


\section{Introduction}

\IEEEPARstart{N}{umber} of IT devices and their computational ability have grown significantly, continuously generating a vast volume of diverse dataset. This fostered the AI technology to become increasingly prospective, which has created and established the values through a variety of utilizations in miscellaneous domains. On account of the enhanced computation performance of the ubiquitous local machines, implementing AI algorithms such as deep learning or machine learning in such devices are now prevalent in order to discover semantics and to create values. In this context, Federated Learning (FL) [1-3] architecture have been suggested, attracting the countless attention in both IT industry and academia, as an apposite framework to be utilized in practical applications, which accommodates the heterogeneity of distributed local dataset. However, acquiring IID (Independently and Identically Distributed) data in a designated local machine is not guaranteed in a real-world distributed environment, as most of the existing systems tend to be biased and skewed due to intrinsic propensity and external factors that influence the circumstances. Thus, measures in order to upgrade the FL performance were introduced to successfully mitigate the unstable impact of aggregating non-IID data through approaches such as utilizing learning paradigms [4-9], clustering [10-14], knowledge distillation [15-18], stochastic approach [11], [19-20], Bayesian theorem [21-24], etc. Initially, FL architecture was specifically designed in order to accommodate the effective training among the distributed conditions without sending the local data itself for security purposes [1], [2], [25], thus many researches were conducted on FL to achieve higher privacy such as detecting adversaries [12], [26]. Moreover, FL has been applied in the network domain as well, in order to construct efficient communication system architecture among heterogeneous local models to reduce the overhead [27], [28]. Likewise, FL was suggested as a viable methodology that effectively encompasses the geographically dispersed local models that were trained independently, which lead the diverse domains to implement the FL architecture such as in healthcare [17], [29-33], IoT [34-37] or Security [38-44].

However, the vanilla FL suffers from the biased and skewed non-IID local dataset, drifting the convergence with low trainability. This seem natural as basic FL model was originally constructed to aggregate the local models that were fully trained with IID dataset. In this research, we suggest a novel label-wise clustering algorithm to gain robust convergence in non-IID input data among the FL network. Before we propound our algorithm, we define the six non-IID training scenarios in perspective of three major entities; individual locals, sum of locals in each global communication round, and every local in sum of all global rounds. We experimentally show that the conventional and well-known FL model (FedAvg) [1] does not approach convergence among such non-IID scenarios [3], [45-48], as FedAvg was suitably designed for IID input dataset. Based on this problem, we introduce the label-wise clustering algorithm in order to gain robust convergence in biased and skewed non-IID input dataset. In FedAvg, after training the local models with their private dataset, the global server randomly collects the parameters of partially selected local models. Choosing random clients is proven to be effective when most of the random sets are independent and identically distributed, whereas most real-world cases do not follow this assumption, having skewed and biased distribution of labels. Therefore, our basic idea is to import the local models with the order of highest variance of the corresponding label set and those approximate the uniform distribution, which will maximize the label coverage. This is a heuristic and powerful approach that can maximize the efficiency of current local data which could guide the global model in FL to optimal convergence. According to this basis, we suggest an evaluation metric that quantitatively assesses the local models before training. Thus, we train the optimal clients only which offers the profound reduction of computational resources, followed by aggregating the parameters of the selected models. we validate our approach through topological analysis and design label-wise clustering algorithm that could maintain the convergence in heterogeneous non-IID and when co-existing with IID local dataset. Through experiments, our algorithm was shown to have an average of 24\% higher accuracy than existing FedAvg and FedSGD [1] in non-IID dataset in given identical global epochs, which ultimately led to faster convergence.

The main contributions that this paper offers are as follows:

\begin{enumerate}
\item  Defining six common non-IID cases among FL framework that easily occurs in real-world applications, adapting their data and analyzing the training performance of the vanilla FL model (FedAvg) as global epoch continues.

\item  Suggesting Label-wise clustering algorithm that efficiently leads to stable convergence through clustering the local models based on their local class labels.

\item  Reducing the computational resources needed to obtain convergence in FL framework by quantitatively evaluating the local dataset which ultimately determines the quality of model in advance to training, simultaneously demonstrating high performance capability in FL through multiple experiments.
\end{enumerate}

This paper is organized as follows. In section II, we enumerate and explain the past studies of obtaining optimizations of FL architecture in non-IID dataset. In section III, we define the six general cases that aggregated the local dataset in three perspectives of overall FL architecture, with corresponding experiments that validates the FedAvg is being trained languidly in such cases. Additionally, we sketch our main approach with topological analysis regarding aggregating multiple local data labels.  In section IV, we construct the label-wise clustering algorithm, delineating the details. Finally, we conclude the paper summarizing our works and future works.

\section{Related Works}

Federated Learning [1-3] is a novel architecture that was recently suggested by Brendon \textit{et al}. [1], which effectively integrates the diversity of locally distributed dataset, as storage capacity and computation ability has substantially been enhanced for local devices in the modern IT era. Nonetheless, when importing local models that were persistently trained with biased and skewed local data, it leads to contamination of global performance and occasionally evanescing the pre-trained weights (\textit{e.g.} client drift [19], [49]) which requires constructive measures to guide the weights to reach the optimal convergence.

Diverse past studies were conducted to decrease, filtrate or prognose the potential effect of non-IID input data. Briggs \textit{et al}. [10] defined the four primary non-IID status, suggesting the appropriate combination of distance and linkage that shows the best performance when implementing the hierarchical clustering among the local clients in Federated Learning through multiple experiments. Chen \textit{et al}. [50] proposed three clustering algorithms in FL and analyzed that clustering in FL acquires faster loss convergence than FedAvg in non-convex optimization, showing that clustering approach in FL comes effective. Ouyang \textit{et al}. [51] implemented dropout mechanism which drops the straggler models that were not well-trained, and selected local models based on the weight correlations. Authors clustered the data after reducing its dimension using PCA, and measured the similarity of local models. Kim \textit{et al}. [52] designed dynamic clustering through GAN, which continuously updates the cluster through latent vectors generated from GAN. Ghosh \textit{et al}. [13] devised an iterative clustering in FL (IFCA), which has shown to be effective in strongly convex functions. Li \textit{et al}. [53] suggested the upgraded algorithm based on the IFCA, which lets the clusters to be overlapped which could congregate the clients in a more successful way, thus achieving better convergence. Xie \textit{et al}. [11] presented a new approach called multi-center FL clustering, by constructing multiple global models and assigning them to the nearest local models. Many studies [17], [54], [55] have implemented auxiliary public dataset to minimize the biased distribution of class labels of existing clients, which accommodates the improvement of training performance in FL. Li \textit{et al}. [16] curated an additional public dataset that could assist the training among the local models and adopted transfer learning with knowledge distillation in the FL network. Yang \textit{et al}. [56] utilized updated model weights to estimate the class distribution utilizing supplementary dataset. Using public dataset as auxiliary method comes practical when existing local dataset are considerably biased, as it supplements the diversity of labels. However, as it requires additional storage capacity which comes as an disadvantage to local models that has low computational power. Also, when large number of class label types are to be classified, volume of benchmark dataset may be large even though local training dataset for each class are minute. Our proposed work is analogous to this approach, based on the fact that quality of training class labels determines the quality of FL. Distance-based clustering such as K-means or Hierarchical clustering in FL form clusters with pairwise distances. Considering that trained weights of deep learning models are permutationally invariant, clustering through pairwise metrics is algorithmically ineffective. However, past studies [10-14], [26], [29], [50-52], [53], [57] have empirically shown that it turned out to be partially successful in converging the global loss function in given feature space, but little logical explanations are specified about the reason behind this phenomenon. Furthermore, most of the past works have clustered the pre-trained weights of the selected local models, whereas the local features may overlap, which is abstruse for the clustering algorithm to form productive clusters. Even for distinct dataset such as MNIST [58], when clustering the set of MNIST itself, the results are partially correct [59] since several data shares similar features with disparate labels, having a lookalike shape in 2-dimensional figures such as number 1 and 7. These shortcomings imply that distance-based local weight clustering is an approach that is intractable to explain, which gives us an inspiration to design an algorithm that clusters through the given local labels. 
Although past several studies has emphasized that biased labels are major defect that perturbs the functionality under FL settings, they suggested alternatives [17], [54-56], [60], [61] such as implementing public dataset. To the best of our knowledge, there were no published articles that selects local models through evaluating their corresponding labels with statistical metrics without using auxiliary dataset on FL framework.

\section{ Problem Definition}
\subsection{ Convergence in FedAvg in Non-IID Cases}

Obtaining optimal convergence in FedAvg is known to be an elusive task, and the main reason behind its obscurity is the incoming local non-IID training dataset due to its heterogeneous circumstances and environment. The vanilla form of FedAvg is as follows:

\begin{equation}
{\mathop{\mathrm{min}}_{M_i\in {\mathbb{R}}^d} \sum_{\forall i}{\frac{n(D_i)}{\bigcup_{\forall i} n(D_i)}\cdot \frac{1}{n(D_i)}\sum_{\forall j}{{\mathbb{L}}_j\mathrm{(}y_i,g(x_i,M_i)\mathrm{)}}}\ } 
\end{equation} 

%식 확인

The objective function (1) indicates the minimization of the loss summation as equation (2) is being iteratively being computed. Let ${\mathbb{L}}_j(\cdot )$ is the loss function of class label index $j$, $g(x_i,M_i)$ is a predicted label that $g:x_i\to {\mathbb{R}}^d$ with ground truth $y_i$, input $x_i$ and $M_i=\{W_i,\ B_i|i\in \mathbb{N}\mathrm{\}}$ with $W_i$ is the set of weights and $B_i$ specifies the set of biases of local model index $i$, and $D_i$ denotes the dataset of model $i$. Following (2) is a general equation that explains the gradient descent in machine learning domain, which leads to the direction of decreasing the $|{\mathbb{L}}_j\mathrm{(}y_i,g(x_i,M_i)\mathrm{)}|$ through updating the elements in $M_i$ independently in ${\mathbb{R}}^d$ space where $\eta $ indicates the learning rate, $t$ is the local epoch and $T$ is the global epoch or some past researches nominated this as a global communication.

\begin{equation}  
M^{\left(T,t+1\right)}_i=M^{\left(T,t\right)}_i-\eta \cdot \frac{\partial }{\partial M^{\left(T,t\right)}_i}\cdot {\mathbb{L}}_j\left(M^{\left(T,t\right)}_i\right) 
\end{equation}

FL is mainly used in domains of supervised learning and should be able to achieve its objective such as classification or regression with high performance using the extracted real-time local training dataset from the designated environment. However, most real-world dataset is skewed, biased and does not guarantee a certain statistical distribution [62], which will degrade the function of the model. In this section, we define six prevalent scenarios that train the non-IID data and show that its convergence is not assured by corresponding experiments. We have focused the scenarios on three perspectives in FL: \textit{individual local clients in a single global epoch}, \textit{all local clients in a single global epoch}, and \textit{all local clients in every global epoch}. We denote the set of labels $L_i$ as $L_i=\{{\ell }_j|1\le j\le n_i,(i,j\mathrm{)}\in \mathbb{N}\mathrm{\}}$ where ${\ell }_j$ indicates the class label of data on index $j$, $c_i$ is the local client in index $i$, and $n_i$ is the total amount of data of $c_i$, $n\left(L_i\right)=n_i$. Note that $i$ is a higher set than $j$, which $\exists i\supset \forall j$ since $j$ is an index of data on local $i$. When non-IID, each case may be ${\ell }_j\mathrm{=}{\ell }_{j^{'}}$ or ${\ell }_j\mathrm{\neq }{\ell }_{j^{'}}$ where $\mathrm{1}\mathrm{\le }(j^{'}\mathrm{\neq }j)\mathrm{\le }n_i\mathrm{,\ }j^{'}\in \mathbb{N}$, for $L_i$ could have one type of label out of $n(\mathcal{L})$ labels, where $\mathcal{L}$ is a set of existing unique labels (\textit{e.g.} in MNIST, $\mathcal{L}=\{0,1,2\dots 9\}$).

We define the six non-IID cases as follows:
\begin{enumerate}
\item[1)] \textit{Case 1-A. }
\textit{At }$\mathrm{\exists }T$\textit{, when }${\ell }_j\in L^{(\exists T,\forall t)}_{\exists i}$\textit{, }$\exists {\ell }_{\left(\forall j,train\right)}$\textit{ are identical}
\item[2)] \textit{Case 1-B. }
\textit{At }$\mathrm{\exists }T$\textit{, when }$\mathrm{(}{\ell }_j,{\widetilde{\ell }}_{\jmath })\in L^{\left(\exists T,\forall t\right)}_{\exists i},$\textit{  }$\exists {\ell }_{\left(\forall j,train\right)}$\textit{ are identical and }$\exists {\widetilde{\ell }}_{\left(\forall \jmath ,train\right)}$\textit{ are unidentical}
\item[3)] \textit{Case 2-A.}
\textit{At }$\mathrm{\exists }T$\textit{, when }${\ell }_j\in L^{(\exists T,\forall t)}_{\forall i}$\textit{, }$\forall {\ell }_{\left(\forall j,train\right)}$\textit{ are identical}
\item[4)] \textit{Case 2-B. }
\textit{At }$\mathrm{\exists }T$\textit{, when }$\left({\ell }_j,{\widetilde{\ell }}_{\jmath }\right)\in L^{(\exists T,\forall t)}_{\forall i},$\textit{ }$\forall {\ell }_{(\forall j,\ train)}$\textit{ are identical and }${\forall \widetilde{\ell }}_{(\forall \jmath ,train)}$\textit{ are unidentical}
\item[5)] \textit{Case 3-A.}
\textit{At }$\mathrm{\forall }T$\textit{, when }${\ell }_j\in L^{(\forall T,\forall t)}_{\forall i}$\textit{, }$\forall {\ell }_{(\forall j,\ train)}$\textit{ are identical}
\item[6)] \textit{Case 3-B. }
\textit{At }$\mathrm{\forall }T$\textit{, when }$\mathrm{(}{\ell }_j,{\widetilde{\ell }}_{\jmath })\in L^{(\forall T,\forall t)}_{\forall i}$\textit{, }$\forall {\ell }_{(\forall j,\ train)}$\textit{ are identical and }$\forall {\widetilde{\ell }}_{(\forall \jmath ,train)}$\textit{ are unidentical}
\end{enumerate}

The perspective of cases change from individual $\exists c^{(\exists T)}_i$ (case 1) to $\forall c^{(\exists T)}_i$(case 2) and to $\forall c^{(\forall T)}_i$ (case 3) where ${\ell }_{(j,\ train)}$ specifies training label of index $j$, and $\exists {\ell }_{\left(\forall j,train\right)}\in {\mathcal{L}}_i\in D_i$. Case 1-A, 1-B are defined by the perspective of an individual local client at $\exists T$. In case 1-A, the given training label $\exists {\ell }_{(\forall j,train)}$ in $\exists c_i$ are identical, which $\exists {\ell }_{(\forall j,train)}$ is composed of a single solitary label, $\exists {\ell }_{\left(j\neq j^{'},train\right)}=\exists {\ell }_{\left(j^{'}\neq j,train\right)}$ and ${\sigma }^2\left(L_{\exists i}\right)=0$, where ${\sigma }^2(x)$ indicates the variance of input x. An important rule when computing ${\sigma }^2$, we consider each label as an independent entity without assigning the numerical correlations or its arithmetic meaning. For instance, label types 0$\mathrm{\sim}$9 are to express their distinctive existence, which is similar to the reason for altering the label into one-hot encoding. Case 1-A considers the worst scenario when only a single type of label is being entered from the $c_i$. In case 1-B however, we set the training labels may be inconsistent and skewed. In this case, we added a small number of randomized labels ${\widetilde{\ell }}_{\jmath }$ along with the large number of unique major label ${\ell }_j$ where $1\le \jmath \neq j\le n_i$ and $\left(\bigcup_{\forall \jmath }{{\widetilde{\ell }}_{\jmath }}\right)\cup \left(\bigcup_{\forall j}{{\ell }_j}\right)=L_i$. For example, when ${\ell }_j$ is an integer label $x$, ${\widetilde{\ell }}_{\jmath }\neq x$, with $n\left(\bigcup_{\forall \jmath }{{\widetilde{\ell }}_{\jmath }}\right)\ll n\left(\bigcup_{\forall j}{{\ell }_j}\right)$ which applies to case 2-B, 3-B as well, where the exact numbers used in the experiment are mentioned in the next subsection. Case 2-A indicates the distributed ${\ell }_j$ in $\forall c_i$ at the equivalent $T$, which contains the one unique type of label per T and $\bigcup^{\ }_{\forall T}{L^{\left(T\right)}_{\left(\forall i,train\right)}}\supset \mathcal{L}$ with $L^{(T)}_{\forall i}\neq L^{(T+1)}_{\forall i}$. Case 2-B appends additional random ${\widetilde{\ell }}_{\jmath }$ with major ${\ell }_j$ in $c_i$, which increases the variance ${\sigma }^2\left(L^{(\forall T)}_{\forall i}\right)$ and $\left(\bigcup_{\forall \jmath }{{\widetilde{\ell }}_{\jmath }}\right)\cup \left(\bigcup_{\forall j}{{\ell }_j}\right)=L_i$ just like case 1-B. When computing ${\sigma }^2(L_i)$, because $L$ indicates a semantic meaning of label, which does not preserve the actual significance of the number, we consider ${\sigma }^2(L)$ as a series of independent objects based on relationships such as $L=\{1,5,10\}$ is equivalent to $\{0,1,2\}$, which gives the invariant statistics. In case 3-A, we randomly allocate the $L^{(\exists T)}_{(i,\ train)}$ which is $\bigcup_{\forall i}{L^{(T)}_{(i,train)}}\supset \mathcal{L}$ or $\bigcup_{\forall i}{L^{(T)}_{(i,train)}}\subset \mathcal{L}$. Finally, case 3-B contains random ${\widetilde{\ell }}_{\jmath }$ with major ${\ell }_j$ in $c_i$, likewise increasing the ${\sigma }^2(L^{\forall T}_{\forall i})$ and $\left(\bigcup_{\forall \jmath }{{\widetilde{\ell }}_{\jmath }}\right)\cup \left(\bigcup_{\forall j}{{\ell }_j}\right)=L_i$. All given cases are biased and skewed, as ${\ell }_j$ are the majority label with single identical label, which implies the worst case of having an imbalanced distribution.

\subsection{ Convergence Result of FedAvg in Six Non-IID Cases}

In order to validate that the FedAvg cannot obtain the optimal convergence in non-IID setting, we empirically test the six cases explained in prior subsection, with randomly selected 30 local clients among 100 client populations to conduct image classifications using MNIST dataset, each client trained with 4 epochs, 32 batch size and 6 layers (Conv2D-Pooling-Conv2D-Flatten-Dense-Dense) using Adam optimizer and categorical cross entropy loss function. Since MNIST is already a well-preprocessed dataset which CNN locates its features smoothly, and local devices in FL may contain low computational processors compared to servers such as laptop, mobile phone or tablet, thus high local epoch is unnecessary. $n(D_{\exists i})=290$, where $D_i$ refers to the unique dataset of $c_i$, in case (1,2,3)-A, having ${\ell }_j\in D_i$ and $n(L_i)=290$, whereas case (1,2,3)-B have $({\ell }_j,\ {\widetilde{\ell }}_j)\in D_i$, $n(\bigcup^{\ }_{\forall j}{{\ell }_j})=200$ and randomly selecting 90 of ${\widetilde{\ell }}_{\jmath }$ among 9 unique ${\widetilde{\ell }}_{train}$, having $n\left(\bigcup^{\ }_{\forall \jmath }{{\widetilde{\ell }}_{\jmath }}\right)=90$. We trained the FedAvg for 30 trials in each case. The accuracy and loss result in given cases are shown in Fig. 1 and 2 (a, b) respectively, wherein Fig. 2 (b) is an amplified version of Fig. 2 (a).
Although $\bigcup^{\ }_T{L^{(T)}_{(\forall i,train)}}\supset \mathcal{L}$ in all cases, results of cases (1,2,3)-A proves that the models were not trained whereas models from cases (1,2,3)-B were partially trained. Moreover, the result of case 1-A was less trained than case (2,3)-A, which indicates that its heterogeneity is more intense, due to ${\sigma }^2_{c1A}\left({\mathcal{L}}^{\left(\forall T\right)}\right)>{\sigma }^2_{c2A,3A}\left({\mathcal{L}}^{\left(\forall T\right)}\right)$ where $c1A$ denotes the case 1-A for sake of brevity, along with other cases. In case 1-A, $\sum_{\forall T}{\frac{n(l_i)}{n({\mathcal{L}}^{(T)}_i)}}\approx \sum_{\forall T}{\frac{n(l_i')}{n({\mathcal{L}}^{(T)}_{i'})}}$ where $l\in \mathcal{L}$ and $i\neq i^{'}$, and will ultimately follow uniform distribution with random label $l\sim U(a,b)$, $f\left(l\right)=\frac{1}{b-a}$ and $a\le \ell \le b$ in each $\exists T$. However, case (2,3)-A does not show the uniform distribution in each $\exists T$, but in $\bigcup^{\ }_{\forall T}{\bigcup_{\forall i}{{\mathcal{L}}^{(T)}_i}}$. It may seem appropriate that case (1,2,3)-B is not trainable, due to their asynchronous labels. However, ${\sigma }^2_{c1B}\left(\bigcup^{\ }_{\forall T}{\bigcup_{\forall i}{{\mathcal{L}}^{(T)}_i}}\right)\approx {\sigma }^2_{c2B}\left(\bigcup^{\ }_{\forall T}{\bigcup_{\forall i}{{\mathcal{L}}^{(T)}_i}}\right)\approx {\sigma }^2_{c3B}\left(\bigcup^{\ }_{\forall T}{\bigcup_{\forall i}{{\mathcal{L}}^{(T)}_i}}\right)$, in fact, ${\sigma }^2_{all\ cases}\left(\bigcup^{\ }_{\forall T}{\bigcup_{\forall i}{{\mathcal{L}}^{(T)}_i}}\right)$ have analogous values in holistic view, which $n(\bigcup^{\ }_{\forall T}{\bigcup_{\forall i}{L^{(T)}_i}})$ in all cases are identical: $261,000(=290\cdot 30\cdot 30)$. 

\begin{figure}
\centerline{\includegraphics[width=19pc]{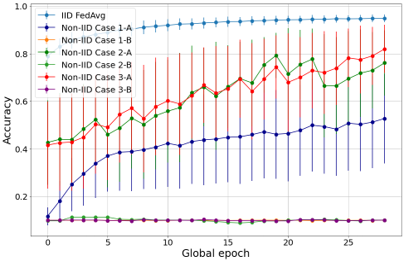}}
\caption{Comparison of FedAvg test data accuracy when training six non-IID data and one IID data with $\widetilde{\sigma}(accuracy_{(T,i)})$.}
\end{figure}

\begin{figure}
\centerline{\includegraphics[width=20.4pc]{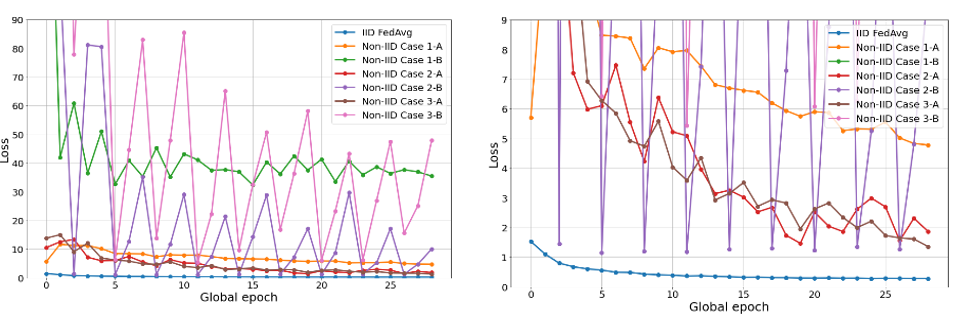}}
\caption{Comparison of FedAvg test data loss when training six non-IID data and one IID data. Due to different impact of training data, the range of y-axis that indicates the loss shows different in (a, b). $\widetilde{\sigma}(loss_{(T,i)})$ was not indicated.}
\end{figure}
Table 1 shows the accuracy and loss after training for 30 global epochs for six non-IID cases and FedAvg trained with the IID dataset. The standard deviation of each metric was calculated as $\widetilde{\sigma }(X)\mathrm = {\sum^{30}_{trial=1}{\sum^{30}_{T=1}{\sum^{30}_{i=1}{\sigma (X_{(T,i)})}}}}\cdot\frac{1}{30\cdot 30\cdot 30}$ where $X$ is an input of accuracy and loss at \textit{T} and local $i$.

\begin{table}
\caption{TEST ACCURACY AND LOSS FOR GIVEN CASES AFTER \textit{T}=30}
\label{table}
\tablefont%
\setlength{\tabcolsep}{12pt}
\begin{center}
\begin{tabular}{| c | c | c |}    %\begin{tabular}{c | c | c } = 중간 선 있음
\Xhline{2\arrayrulewidth}
Cases & Accuracy (\%) & Loss \\ [0.75ex] 
\Xhline{2\arrayrulewidth}
FedAvg IID & 94.83 ($\pm2.34$) & 0.29 ($\pm0.23$) \\ 

Case 1-A Non-IID & 52.77 ($\pm17.16$) & 4.79 ($\pm8.83$) \\

Case 1-B Non-IID & 10.09 ($\pm0.57$) & 35.54 ($\pm35.54$)  \\

Case 2-A Non-IID & 76.15 ($\pm18.52$) & 1.87 ($\pm5.33$)  \\

Case 2-B Non-IID & 10.11 ($\pm0.65$) & 10.09 ($\pm1.16$)\\ 

Case 3-A Non-IID & 81.99 ($\pm17.89$) & 1.36 ($\pm5.67$) \\ 

Case 3-B Non-IID & 10.10 ($\pm0.54$) & 47.95 ($\pm1.16$) \\ [0.75ex] 
\Xhline{2\arrayrulewidth}
\end{tabular}
\end{center}
%\label{tab1}
\end{table}

\section{ Label-wise Clustering}

\subsection{ Label-wise Clustering}

In order to minimize the heterogeneity and biased distribution of the labels in a given environment, we implement the label-wise clustering to overcome the frozen or untrainable weights. As we have shown in section III.1, when the variance of the set of unique labels among all the models in synchronous global epoch is high, it triggers the highest training performance, which implies that optimal convergence in FL can be gained through training various labels of dataset. Since FL is implemented in supervised learning [2], [3], we assume that the input dataset is preprocessed and labeled in advance, thus we suggest the clustering method to guarantee the acquisition of fairly distributed labels in non-IID local training data. As we have shown that the input local dataset is skewed toward certain label(s), this would lag the weights to be inactive. In such a scenario, our approach for this problem is to cluster the locally-trained models in advance and select only the locals which could guarantee the convergence.

When FL randomly selects the $n$ models among $S_T$, where we denote the set of local clients at $T$ as $S_T=\left\{c_i\mathrel{\left|\vphantom{c_i 1\le i\le N,\ i\in \mathbb{N}}\right.\kern-\nulldelimiterspace}1\le i\le N,\ i\in \mathbb{N}\right\}$, $n(S_T)=N$. Also, we nominate the set of selected random clients as $s_T=\{{\tilde{c}}_h|1\le h\le \mathcal{N},\mathcal{N}\le N,\ h\in \mathbb{N}\mathrm{\}}$ where ${\tilde{c}}_h$ is the chosen client of sequential index $h$, which ${\tilde{c}}_h$'s dataset ${\tilde{D}}_h\ni {\widetilde{\ell }}_{\jmath }$  and $s_T\subset S_T$. We compute the label-wise clustering ${\mathcal{C}}_k$, where ${\mathcal{C}}_k=\left\{M_i\right|1\le i\le N,\ 1\le k\le \mathcal{L},\ (i,k\mathrm{)}\in \mathbb{N}\mathrm{\}}$, $k$ indicates the index of the clusters and $M$ is the local model, forming $n\left({\mathcal{C}}^{(T)}_k\right)=K^{(T)}=n({\mathcal{L}}^{(T)})$ that refers to the number of clusters $K$ in $T$ is equivalent to the set of unique labels in $\mathcal{L}$ in corresponding $T$. Recall that in heterogeneous settings, ${\mathcal{L}}^{(T)}={\mathcal{L}}^{(T^{'})}$ or ${\mathcal{L}}^{(T)}\neq {\mathcal{L}}^{(T^{'})}$ and ${\mathcal{C}}^{(T)}_k\cap {\mathcal{C}}^{\left(T\right)}_{k^{'}}\neq \emptyset $ or ${\mathcal{C}}^{(T)}_k\cap {\mathcal{C}}^{\left(T\right)}_{k^{'}}=\emptyset $ where $T\neq T^{'}$ and $k\neq k^{'}$. Fig. 3 represents the idea pictorially, when $k=3$, ${\mathcal{C}}^{(T)}_k\cap {\mathcal{C}}^{\left(T\right)}_{k^{'}}\neq \emptyset $. We denote the set of intersection areas as $A=\{A_p|1\le p\le q,\ \left(p,q\right)\in \mathbb{N}\mathrm{\}}$\textit{, }where $A_p$ specifies the vicinity covered by $p$-number of clusters, $A_p=\bigcup_{\forall k}{({\mathcal{C}}_k-\bigcup^q_{\hat{p}=2}{A_{\hat{p}}})}$\textit{, }$A_1=\bigcup_{\forall k}{{\mathcal{C}}_k}-\bigcup^q_{\hat{p}=2}{A_{\hat{p}}}$\textit{, }$q$\textit{ }indicates the\textit{ }number of labels and $p$ is the index. Fig. 3 displays the topology of three clusters that encompass the given models trained with three total labels (\textit{e.g.} 0,1,2), forming ${\mathcal{C}}_p$ where $\mathrm{1}\le p\le 3$. 

\begin{figure}
\centerline{\includegraphics[width=18.5pc]{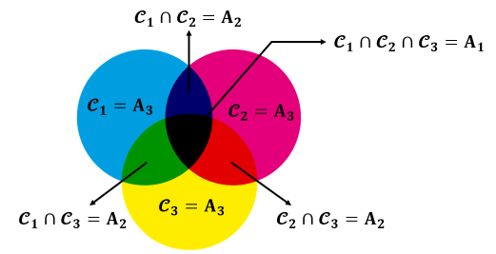}}
\caption{Clusters encompassing the local models trained with heterogeneous labels (3 types of class labels) and their designated areas in given FL network.}
\end{figure}

To elaborate, ${\mathcal{C}}_1$ covers $c_i$ that were trained with ${\mathcal{L}}_i$ that covers corresponding labels (\textit{e.g.} $\mathrm{1}\in {\mathcal{L}}_i$). Moreover, $A_3$ contains only the local models trained with a single type of label, whereas $A_2$ holds models trained with two labels, and ${\mathrm{A}}_{\mathrm{1}}$ encompasses models that were trained with all three labels. To generalize, ${\mathcal{C}}_p\ni c_i$ when defining $c_i$ with objective function (1), updating with (2). The Fig. 4 indicates the distinct area covered by $\exists {\mathcal{C}}_k$, where $1\le k\le 10$, $k\in \mathbb{N}$ having ${\sigma }^{2}\left(A_6\right)>{\sigma }^{2}\left(A_{7\le p\le 10}\right)$, which $A_{1\le p\le 5}=\emptyset $, $p_{min}=6\ $.

\begin{figure}
\centerline{\includegraphics[width=17.5pc]{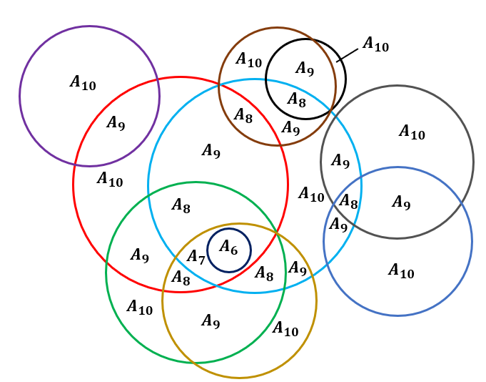}}
\caption{Example of distributed clusters in FL network when 10 types of class labels exist.}
\end{figure}

%여기부터 ................ so far so good

When given numerous local models trained with multiple labels, we compute the ${\mathcal{C}}_k$ and $A_k$ based on the intersections of input class labels. This is feasible due to the fact that $L^{(\exists T)}_{(i,\ train)}$ are pre-labeled in advance, and $L^{(\exists T)}_{(i,\ test)}$ are already predicted before the phase of aggregation. After the label-wise clustering, we select the ${\tilde{c}}_h\in s_T$ with the priority of ${\tilde{c}}_h$ which follows the sequential condition: $A_{1}>A_{2}>\dots >A_{n(\mathcal{L})-1}$, since ${\sigma }^{2}\left(L^{\left(T\right)}_{(A_1,i)}\right)>{\sigma }^{2}(L^{(T)}_{(A_{p>1},i)})$, having ${\mathcal{L}}_{A_1}>{\mathcal{L}}_{A_{p>1}}$ and locating the index $p$ that $arg{\mathop{\mathrm{max}}_{1\le p\le n({\mathcal{L}}_i)\ } {\sigma }^2(L^{(T)}_i)\ }$. $A_p$ is a set composed of $c_i$ which could be $D^{(T)}_i\neq D^{(T)}_{i^{'}}$ and ${\mathcal{L}}^{(T)}_i\neq {\mathcal{L}}^{(T)}_{i^{'}}$ where $i\neq i^{'}$. Under condition that $\forall c_i\in A_p$ are trained with the corresponding ${\mathcal{L}}_i\mathrm{\in }D_i$, we suggest the following equation (3) and attain $s_T$ after $\mathcal{N}$ iterations, where $R$ specifies the random selection, ${\left(x\right)}_{\mathcal{N}}$ is the iteration of input $x$ for $\mathcal{N}$ times and $S_T[i]$ indicates the element of $S_T$ in index $i$, which is equivalent to $c_i$. 

\begin{equation}
{R:\exists c_i\to {\tilde{c}}_h} \; \textit{s.t.} \, {\left(h=arg{\mathop{\mathrm{max}}_{\forall i} \frac{{\sigma }^2\left(S_T\left[i\right]\right)}{n_i}}\ then\mathrm{\ }s_T\oplus {\tilde{c}}^{\left(T\right)}_h\right)}_{\mathcal{N}}
\end{equation}
%eqref{GrindEQ__3_} (3) 뒤에도 이거 붙이면 바로 갈 수 있음
% $$ 이거 빼야되는 데 빼면 (3)이 내려감

The initial $s_T$ is empty (\textit{e.g. }$s_{T=0}=\emptyset $) and whenever iteration of (3) is  being computed, it simultaneously deletes the corresponding $c^{(T)}_i={\tilde{c}}^{\left(T\right)}_h$ in $S_T$ where $\oplus $ denotes the concatenation and $\frac{{\sigma }^2(c_i)}{n_i}$ is to measure the variance considering the number of elements. The assorted result of $s_T$ is equivalent with $A_p\oplus A_{p+1}\oplus \dots \oplus A_{p+\mathrm{c}}$ where integer $\mathrm{c}$ ranges $0\le \mathrm{c}\le (\mathrm{n(}\mathcal{L})-\mathrm{1)}$, $\mathrm{c}\in \mathbb{Z}$. ${\left(s_T\oplus {\tilde{c}}^{\left(T\right)}_h\right)}_{\mathcal{N}}\mathrm = s_T$, as it concatenates ${\tilde{c}}^{\left(T\right)}_h$ to existing $s_T$, and we determine $s_T$ as our final input to be aggregated in the global server. 

\subsection{Arithmetic Relation through Dissecting Clusters}

In this section, we search the relationships that defines the numerical properties of the areas and clusters among the given FL network.  In our label-wise clustering, $\bigcup^{\ n({\mathcal{L}}_i)}_{k=1}{{\mathcal{C}}_k}\supseteq \forall c_i$ that were trained with (1), (2) using ${\mathcal{L}}_i\in D_i$. Among $\bigcup^{\ n({\mathcal{L}}_i)}_{k=1}{{\mathcal{C}}_k}$\textit{, }$A_{p_{min}}$ has an area that is $A_{p_{min}}=\bigcap^{\ n({\mathcal{L}}_i)}_{k=1}{{\mathcal{C}}_k}$ where ${\mathcal{C}}_{\forall k}$ intersects and $1\le p_{min}\le n({\mathcal{L}}_i)$. We define the total number of $A^{(T)}$ in order to efficiently designate the components by $A^{(T)}_p$. Let ${\tau }^{(T)}_i=\{n\left({\mathcal{L}}^{(T)}_i\right)|1\le i\le N\}$, and number of $A$ at \textit{T}, $n\left(A^{(T)}\right)$ increase in non-linear fashion as $n({\mathcal{L}}_i)$ enlarges linearly. The upper bound of $n\left(A^{(T)}\right)$ can be defined as (4).

\begin{equation}
{\mathrm{sup} (n\left(A^{(T)}\right))\ }=\mathcal{F}({\tau }^{\left(T\right)}_i)\mathrm{\stackrel{def}{=}}1+{\tau }^{(T)}_i\cdot ({\tau }^{(T)}_i-1) 
\end{equation}

When multiplying out the terms of (4), $\mathcal{F}({\tau }^{\left(T\right)}_i)={\left({\tau }^{\left(T\right)}_i\right)}^2-{\tau }^{(T)}_i+1$, and we prove $\mathcal{F}({\tau }^{(T)}_i)$ through proof of induction. In the basis case, when ${\mathcal{L}}^{(T)}_i$ contains only one type of unique label, ${\tau }^{(T)}_i=1-1+1=1$ where existing $A^{(T)}_p$ has a sole area; $A^{(T)}_{p_{min}}$, which is true. In the induction step where $2\le {\tau }^{(T)}_i$, $\mathcal{F}\left(2\right)=1+2$, and as ${\tau }^{(T)}_i$ linearly increases by one, such as $\mathcal{F}\left(3\right)=1+3+3$ and $\mathcal{F}\left(4\right)=1+4+4+4$, which validates $\mathcal{F}({\tau }^{(T)}_i)$ is true. When $\mathcal{F}\left({\tau }^{(T)}_i+1\right)={\left({\tau }^{(T)}_i+1\right)}^2-\left({\tau }^{(T)}_i+1\right)+1=\mathcal{F}({\tau }^{(T)}_i)$, this is also true. $\mathcal{F}\left({\tau }_i\right)=1+{\tau }^{(T)}_i+{\tau }^{(T)}_i+\dots +{\tau }^{\left(T\right)}_i=1+\sum^{{\tau }^{\left(T\right)}_i-1}_{i=1}{{\tau }^{\left(T\right)}_i}$, the initial term integer $1=n\left(A^{\left(T\right)}_{p_{min}}\right)=n(\bigcup_{\forall p}{A^{(T)}_p}-\bigcup_{p_{min}<p\le n({\mathcal{L}}_i)}{A^{(T)}_p})$ and the second term (\textit{i.e.} initial ${\tau }^{(T)}_{i=1}$) denotes $n\left(A^{\left(T\right)}_{p_{min}+1}\right)=n(\bigcup_{\forall p}{A^{\left(T\right)}_p}-\bigcup_{p_{min}+1<p\le n\left({\mathcal{L}}_i\right)}{A^{\left(T\right)}_p}-A^{\left(T\right)}_{p_{min}})$ and for the third term (\textit{i.e.} ${\tau }^{(T)}_{i=2}$), $n\left(A_{p_{min}+2}\right)=n(\bigcup_{\forall p}{A^{\left(T\right)}_p}-\bigcup_{p_{min}+2<p\le n\left({\mathcal{L}}_i\right)}{A^{\left(T\right)}_p}-A^{\left(T\right)}_{p_{min}}-A^{\left(T\right)}_{p_{min+1}})$. Consecutively, the final term ${\tau }^{(T)}_i$ indicates the $n\left(A_{n({\mathcal{L}}_i)}\right)=n(\bigcup_{\forall p}{A^{\left(T\right)}_p}-\bigcup_{p_{min}\le p<n\left({\mathcal{L}}_i\right)}{A^{\left(T\right)}_p})$. Since the lower bound of $p_{min}$ is 1,  $\mathcal{F}({\tau }^{(T)}_{p_{min}\le i})$ is the upper bound of possible $n\left(A^{(T)}\right)$.

\subsection{ Quantitative Evaluation of Local Models through Dataset without Training}

Each identically layer-structured local model among FL network that are trained through heterogenous local training dataset can be assessed before it trains its model through evaluating their corresponding dataset. Recall that FL is supervised learning, in which its class labels are predefined, giving us the possibility of collecting the information of local models as training AI models require a comprehensive resource consuming process. Since label-wise clustering aims to cluster the local models to sort the clients that encompass the dataset with various distributions before training them. Based on those ${\tilde{c}}^{\left(T\right)}_h\in s_T$, we evaluate their model quality by comparing their distribution. The effect of $s_T$ was experimentally proven partially by six cases in section III-1, where the accuracy of case 1-A was evidently lower than case 2-A and 3-A, and we assume that uniform distribution of training labels encourages optimal and swift convergence. 

We know that ${\sigma }^{\mathrm{2}}\left({\tilde{D}}_h\right)\ge {\sigma }^{\mathrm{2}}\left(D_{i\neq h}\right)$ and implementing the loss function $\mathbb{L}\mathrm{(}{\widetilde{\ell }}_{\jmath },(x_{\jmath },M_{\jmath }))$ such as cross entropy-based loss or mean square error where ${\tilde{D}}_h\ni {\widetilde{\ell }}_{\jmath }$, we compare the ${\widetilde{\ell }}_{\forall \jmath }$ and $(x_{\forall \jmath },M_{\forall \jmath })$. In $\mathbb{L}\mathrm{(}{\ell }_j,(x_j,M_j))$, when ${\exists \ell }_{(j,train)}\mathrm{\in }{\mathcal{L}}_{(j,test)}$ it would occur to have proper training, whereas ${\exists \ell }_{(j,train)}\mathrm{\notin }{\mathcal{L}}_{(j,test)}$ will cause overfitting since features of classes that $\left({\mathcal{L}}_{\left(j,test\right)}\mathrm{-}{\mathcal{L}}_{\left(j,train\right)}\right)\mathrm{\ni }{\exists \ell }_{(j,train)}$ were neglected to tune the $M_j$, which is highly prone to becoming a biased model. Furthermore, since FL framework updates their model in cumulative-basis in sequential \textit{t }with (2) and in \textit{T} (1), single biased training would perturb the framework and adding that contaminated model when averaging at global server will ruin the entire function. Consider four distribution function samples that describes of $L_{1\le i\le 4}$; $p\left(L_1\right),p\left(L_2\right),p\left(L_3\right),p\left(L_4\right)$ where $p(\cdot )$ is the probability density function graphs which are shown in Fig. 5.

\begin{figure}
\centerline{\includegraphics[width=18.5pc]{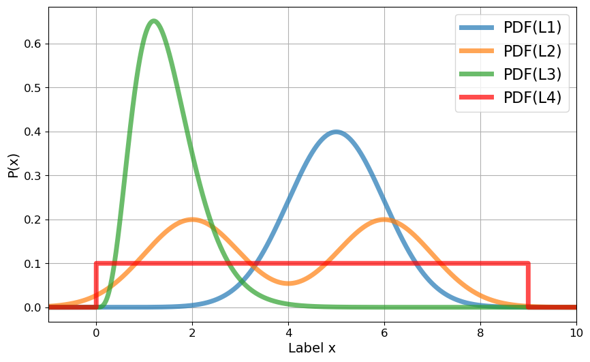}}
\caption{Four sample distribution visualization of $L_1, L_2, L_3, L_4$. Each distribution displays the density of existing labels (\textit{e.g.} PDF(L2) possesses a comparatively large number of labels 2 and 6).}
\end{figure}

Given $p\left(L_{1\sim 4}\right)$ were intentionally set to explicit distribution, normal distribution $\ell \mathrm{\sim }N(5,1)$, mixture distribution which $p\left(L_2\right)=0.5\cdot \left(\ell \sim N\left(2,1\right)\right)+0.5\cdot \left(\ell \sim N\left(6,1\right)\right)$, gamma distribution $\ell \mathrm{\sim }\gamma \mathrm{(5)}$ and uniform distribution $\ell \sim U(0,9)$, respectively. Note that values from x-axis are labels which only stands for definite semantics of given classes, but the numerical parameters are converted values as sequential indexes, which can be statistically calculated. Among four distributions, the model trained with $L_4$ is most likely to outperform the models trained with others due to ${\sigma }^{2}(L_4)$ has the largest value as we have explicated in section III and IV. This concept is similar to supervised learning, as predetermining that specific $L_{\exists i}$ is possible, and will lead to best performance among $L_{\forall i}$, and our aim is to measure how much dissimilarity it contains between our best option which gives us an intuition that a given model can be evaluated without training the neural network. The comparison is computed by using the Kullback Leibler Divergence (5).

\begin{equation} 
KL(p\left(L_i\right)\,||\;p\left(L_{i^{'}}\right))\; \mathrm{\stackrel{def}{=}} \; \sum_{\forall i}({p\left({\ell }_i\right)}\cdot{log \frac{p({\ell }_i)}{p({\ell }_j)}})
\end{equation} 

%% Reason I used KLD : To prove that lowest distribution works in FL, using KLD.

The left term $p\left(L_i\right)\approx \ell \sim U(a,b)$ and index $i^{'}\neq i$. For example, when assigning each $p(\cdot )$ to corresponding places,  $p\left(L_4\right)\mathrm{=}\ell \sim U\left(a,b\right)\mathrm{=}\frac{\mathrm{1}}{b-a}$ (if $a\le \ell \le b$) otherwise 0, and $p\left(L_1\right)=\ell \mathrm{\sim }N\left(\mu ,\sigma \right)=\frac{1}{\sqrt{2\pi }\sigma }e^{\frac{{\left(\ell -\mu \right)}^2}{2{\sigma }^2}}$, $KL(p\left(L_4\right)||p\left(L_1\right))$ = $\sum_{\forall j}{(\frac{\mathrm{1}}{b-a}log\frac{1}{\sqrt{2\pi }\sigma }e^{\frac{{\left(\ell -\mu \right)}^2}{2{\sigma }^2}}-{log \frac{1}{b-a}\ })}$. For $KL(p\left(L_4\right)||p\left(L_2\right))$, let  $\sum_{\forall j}{(\frac{\mathrm{1}}{b-a}log\ (\psi \left(w,\mu ,\sigma \right)+\psi \left(w',\mu ',\sigma '\right))-{log \frac{1}{b-a}\ })}$ where $\psi \left(w,\mu ,\sigma \right)\mathrm{\stackrel{def}{=}}w\cdot \frac{1}{\sqrt{2\pi }\sigma }e^{\frac{{\left(\ell -\mu \right)}^2}{2{\sigma }^2}}$. Likewise, $KL(p\left(L_4\right)||p\left(L_3\right))$ is computed through gamma distribution compared with given uniform distribution. The outcome of $KL(p\left(L_4\right)||p\left(L_{1\le i\le 3}\right))$ are 2093, 602, and 3204 when the base of logarithm is 10.

Since $KL\left({\tilde{L}}_h||U(L_i)\right)\le KL\left(L_i||U(L_i)\right)$, it implies that the distribution of a given client's training label can be quantitatively evaluated by the criteria of $U(L_i)$ where $U(L_i)\approx {\ell }_{\forall j}\sim U(a,b)$ and ${\tilde{L}}_h\mathrm{\neq }L_i$. Likewise, through locating the ${\tilde{L}}_h\ni {\tilde{c}}_h$ that approximates to uniform distribution and comparing with the current distribution $L_i$, the quality of individual local models can be assessed. Therefore, we define proposition and theorem based on the analysis.
\\\\
\textbf{Proposition: } When $\ell \mathrm{\sim } D(a,b) = \ell \mathrm{\sim} U(0,\sigma^{2}(L))$, $ lim_{T \to Y} ( lim_{t \to X} {\mathbb{L}_{(T,t)}\mathrm{(}{\ell }_j,(x_j,M_j))} ) \to 0$, where $(X, Y) > 0$ and $(X,Y) \in \mathbb{N}$.  
% When uniform distribution, loss converges to 0
\\\\
\textbf{Proof}: In (1), $g: x_{(1,i)} \to \mathbb{R}^{d_{1}}$ and $g: x_{(2,i)} \to \mathbb{R}^{d_{2}}$, since $d_{1} > d_{2}$ if corresponding class label set $L_{1}, L_{2}$ exist having $\sigma^{2}(L_{1}) > \sigma^{2}(L_{2})$, (2) adaptively finetunes the $M_{i}$ that approaches to minimization of loss.
\\\\
\textbf{Theorem: } When $\ell \mathrm{\sim } D(a,b) \approx \ell \mathrm{\sim} U(0,\sigma^{2}(L))$, the loss converges having $ lim_{T \to Y} ( lim_{t \to X} {\mathbb{L}_{(T,t)}\mathrm{(}{\ell }_j,(x_j,M_j))} ) \to 0$, where $(X, Y) > 0$ and $(X,Y) \in \mathbb{N}$.  
%% when other distribution is similar to Uniform Distribution => Better Convergence
\\\\
\textbf{Proof: } Based on (5), if $KL(D_{1}(a,b)||U(0,\sigma^{2}(L)))$ $>$ $KL(D_{2}(a,b)||U(0,\sigma^{2}(L)))$ where $L$ is the set of all existing class labels, $ lim_{T \to Y} ( lim_{t \to X} {\mathbb{L}_{(T,t)}\mathrm{(}{\ell },(D_{2},M))} ) > lim_{T \to Y} ( lim_{t \to X} {\mathbb{L}_{(T,t)}\mathrm{(}{\ell },(D_{1},M))} )$, having $\vartriangle M_{(1,j,j+1)}$ $>$ $\vartriangle M_{(2,j,j+1)}$ where $\vartriangle M_{(c,j,j+1)}$ denotes the differentiation of set of parameters between index $j$ and $j+1$ , $M=\bigcup{(W,B)}$ at $D_{c}$.
\\\\

\section{Designing Algorithm}

We propose the methodology of our algorithm based on the implementation of preliminary theoretical analysis in section IV. Apart from theoretical validation, our consideration also lies on the systematic viability inside the FL framework in order to efficiently operate the algorithm. When $\forall c^{(t,T)}_i$ train their corresponding model through (2) for $t$, updating $M_i$, and $\forall c^{(t,T)}_i$ transmit ${\sigma }^2(L^{\left(T\right)}_i)$ to the server in a synchronized manner. Server conducts a sorting algorithm for the set of received ${\sigma }^2(L^{\left(T\right)}_i)$, selects the highest $n$-clients, and responds to each client to direct their $M_i$. We add supplementary condition that ${\mathcal{L}}^{(T)}_h$ are to satisfy  ${\sigma }^2({\mathcal{L}}^{\left(T\right)}_i)\neq 0$, which ${\mathcal{L}}^{(T)}_h\in {\tilde{c}}^{(T)}_h$ and guarantees ${\mathcal{L}}^{(T)}_h\ge 2$. This would incur to increase the trainability since $A^{\left(T\right)}_{p_{min}}$ of ${\sigma }^2\mathrm{(}{\mathcal{L}}^{(T)}_h\mathrm{)}$ satisfies ${\sigma }^2\left({\mathcal{L}}^{\left(T\right)}_h\right)\mathrm{\ge }{\sigma }^2\left({\mathcal{L}}^{\left(T\right)}_{i\neq h}\right)$, which covers most of the existing labels which are to be trained on each $T$ where ${\mathcal{L}}^{(T)}_h\ni L^{(T)}_i\ni {\tilde{c}}_h\ni A^{\left(T\right)}_{p_{min}}$ and ${\mathcal{L}}^{\left(T\right)}_i\mathrm{\notin }A^{\left(T\right)}_{p_{min}}$. Server aggregates the recollected $M^{\left(T\right)}_h\in {\tilde{c}}_h$, propagates $\frac{1}{n}\sum^n_{h=1}{M^{(T+1)}_h}$ to $\forall c_i$, these steps are iteratively performed for an amount of $T$. The overall process is illustrated in algorithm 1, which is operated among the global server and local clients. Compared to previously suggested FL clustering methodology [10-12], [29], [51], [52], [57] that implemented a pairwise distance metric among $M^{(T)}_i$, which requires $O({\mathfrak{n}}^2)$ or $O({\mathfrak{n}}^3)$ computational time complexity when clustering where $\mathfrak{n}$ is the size of input arrays $M_i$. Our algorithm requires $O(n(L^{(T)}_i))$ or $O(n)$ at the best case or $O(n^2)$ in the worst case of time complexity where $n$ is the number of selected models, and in most cases $n<\mathfrak{n}$, reducing the effort by selecting only the local models that were estimated to enhance the overall performance of FL effectively.

\begin{algorithm}
    \caption{Label-wise Clustering in FL Aggregation.}
    \label{alg:algorithm-label}
    \begin{algorithmic}\\
    \textbf{Input:} Set of $D_{(i,train)}$, global epoch $\mathbb{T}$, local epoch $\mathrm{t}$, total number of local clients $N$, number of selected local clients $n$\\
    \textbf{Output:} updated parameters $M$
    
    \State Initialize empty list $LabelSortLst$\textit{, }$LabelSortLstIdx$
    \State Initialize integer $count\gets 0$  
    \For {$T$ = 1, {\dots},$\ \mathbb{T}$} 
        \For {each $c_i$} \textbf{in parallel} 
        \If {${\sigma }^{2}\left(L^{\left(T\right)}_i\right)\mathrm{\neq }\mathrm{0}$} transmit ${\sigma }^{2}\left(L^{\left(T\right)}_i\right)$ to server
        \EndIf
        \If{server received = True} $count++$
        \EndIf
        \If {$count<n$} $n\mathrm = count$  
        \EndIf
        \EndFor
        \For {$i$\textit{ = 1, {\dots}, }$n$}
            \State $LabelSortLst\ \bigoplus \ {\sigma }^2\left(L^{\left(T\right)}_i\right)$ 
        \EndFor
        \State $LabelSortLstIdx\gets argsort(LabelSortLst)$ 
        \For {$i$ = 1, {\dots}, $n(LabelSortLstIdx\mathrm{)}$} \Comment{Server} 
            \State \textit{h} $\gets $ $LabelSortLstIdx[i]$ 
            \State transmit \textit{True} to ${\tilde{c}}^{(T)}_h$ 
        \EndFor 
        \For {$c^{(T)}_i$} \textbf{in parallel} 
            \If {$c^{(T)}_i$ received = \textit{True}} 
                \For {$t$ = 1, {\dots}, $\mathrm{t}$}  
                    \State $M^{\left(T,t+1\right)}_i=M^{\left(T,t\right)}_i-\eta \cdot \frac{\partial }{\partial M^{\left(T,t\right)}_i}\cdot {\mathbb{L}}_j\left(M^{\left(T,t\right)}_i\right)$ 
                \EndFor
                \State transmit $M^{(T)}_i$ to server\textbf{} 
            \EndIf
        \EndFor
        \State receive $M^{(T)}_{\forall i}$ at the server 
        \State \textbf{broadcast} $\frac{1}{n}\sum^n_{h=1}{M^{(T+1)}_h}$ to $\forall c^{(T)}_i$ 
    \EndFor 
    \end{algorithmic}
\end{algorithm}

At algorithm line 17, when $c^{(T)}_i$ had received a consent from the server that its $M^{(T)}_i$ is for collection, it starts training their parameters and saving other unselected client's resources, reducing $O(\forall i-\forall h)$ computation. Similar to FedAvg, clients send their $M^{(T)}_i$ to the server, and server aggregates the parameters and broadcasts it back to each client iteratively updating their parameters until global epoch expires.

\section{Experiments}

In this section, we experimentally demonstrate the efficiency of our algorithm and compare its accuracy with other FL algorithms when training non-IID local dataset as global epoch continues. We demonstrate three types of experiments to validate the effect of our designed algorithm. The entities in every experiment has are consisted of one global server and 100 local clients ($c_{1\le i\le 100}$), and randomly select 30 clients; $n\left(s_{\forall T}\right)=30$. Every client carries an identically structured with the equivalent hyperparameter values of CNN model to classify the given image data, with the layer structure of Conv2D -- MaxPooling -- Con2D -- MaxPooling -- Flatten -- Dense -- Dense. Specified structure of the model is set with a comparatively small number of layers to indicate the low computation ability of local clients such as mobile devices to operate the lightweight AI models. In addition, the global epoch was set to 30, local epoch to 4 with a batch size of 32 in FedAvg and Label-wise Cluster FL.
The first experiment is to measure and compare the accuracy outcome of three FL algorithms: FedAvg, FedSGD and Label-Clustering FL with a local dataset that has biased distribution. For the dataset, we used Fashion MNIST (FMNIST) [63] and Cifar-10 [64]. We set three scenarios each with a different probability $p(x_i)$ of having biased non-IID dataset: 0.7 (70\%), 0.4 (40\%) and 0.1 (10\%) where $x_i$ is an event of being trained with biased non-IID data $D_i$. To signify the worst case, ${\exists D}_i$ were set with one unique label type where $n\mathrm{(}{\mathcal{L}}_i)=1$. For other $i$, each scenario with $1-p\left(x_i\right)=$ 0.3, 0.6 and 0.9, and their ${\mathcal{L}}_i$ were randomly selected among the population of 60,000 training dataset in FMNIST and 50,000 training dataset in Cifar-10. Test dataset were identically used in $\mathrm{\forall }T$. The number of local data $n_i$ of $\forall c_i$ were also randomly selected, between the range of $30\le n_i\le 270$, $\mu \left(n_i\right)\approx 150$ having lower bound with 30, since $n\left(s_T\right)=30$. The accuracy results of FMNIST and Cifar-10 are presented in Fig. 6 and 7 respectively, which shows the average values of 30 trials. 

In Fig. 6 and 7, the red, blue, and green lines respectively specify the training result of FedAvg, FedSGD and Label-wise Clustering. Also, the graphs contain three different types of data points: square, triangle and circle, each indicating the simulation result of $p\left(x_i\right)=$ 0.7, 0.4 and 0.1. When $p\left(x_i\right)=0.7$, randomly selected ${\tilde{c}}_h$ shows that it has lower accuracy achievement in FedAvg and FedSGD, whereas the label-wise clustering maintains stable trainability in such non-IID status.

The second experiment implemented the identical conditions in section III-1 case (1,2,3)-A, which displays the training performance of FedAvg in each non-IID case, each having approximately 55.6\%, 62.8\% and 77.5\% of accuracy. We adopted our algorithm into these cases, increasing the corresponding accuracy to 72.4\%, 74.5\% and 93.2\% as suggested in Fig. 8 along with stably diminishing loss in Fig. 9. In Fig. 8 and 9, the green lines refer to the outcome of the label-wise clustering algorithm where the red lines are the result of FedAvg. The circular data point in each line denotes case 1-A, triangular point is case 2-A, and square point specifies the 3-A.

Our final experiment is to demonstrate the impact of the proportion of IID and non-IID label dataset among each $D_{\exists i}$. Fig. 10 indicates the average of final accuracy of 30 selected local models when $0.1\le p\left(x_i\right)\le 0.9$, and overall FL is trained with FedAvg (orange bar graph) and Label-wise Clustering (blue bar graph). The black line displays the range of standard deviation of accuracy within 30 trials. Following experiment was conducted with identical settings in Fig. 8 and 9, except that the global epoch was set to 20. The result shows that when the proportions of IID increases, the accuracy tends to increase linearly in FedAvg. However, our Label-wise Clustering remains stable, having around 93.8\% accuracy in every proportion in FMNIST and 37.6\% in CIFAR-10. Based on this outcome, we can refer that performance of FedAvg is directly affected by the proportion of non-IID dataset.

\begin{figure}
\centerline{\includegraphics[width=18.5pc]{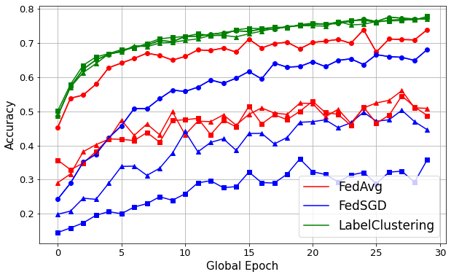}}
\caption{Accuracy results of three FL algorithms when training non-IID FMNIST dataset in given FL network.}
\end{figure}

\begin{figure}
\centerline{\includegraphics[width=18.7pc]{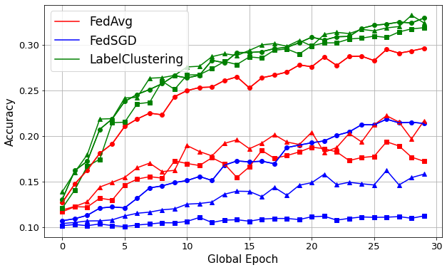}}
\caption{Accuracy results of three FL algorithms when training non-IID Cifar-10 dataset in given FL network.}
\end{figure}

\begin{figure}
\centerline{\includegraphics[width=18.5pc]{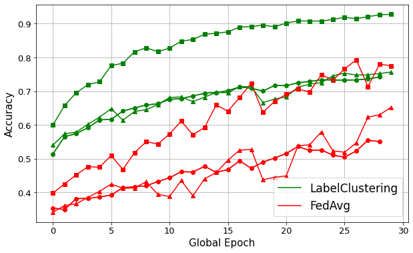}}
\caption{Accuracy result visualizations of two FL algorithms when training three non-IID cases (1,2,3 - A) in given FL network.}
\end{figure}

\begin{figure}
\centerline{\includegraphics[width=18.5pc]{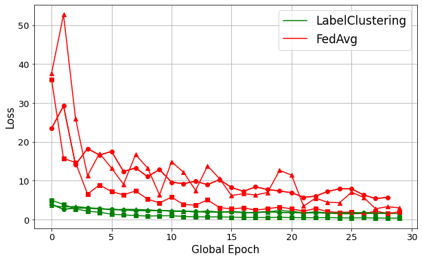}}
\caption{Loss result visualizations of two FL algorithms when training three non-IID cases (1,2,3 - A) in given FL network.}
\end{figure}

\begin{figure}
\centerline{\includegraphics[width=18.5pc]{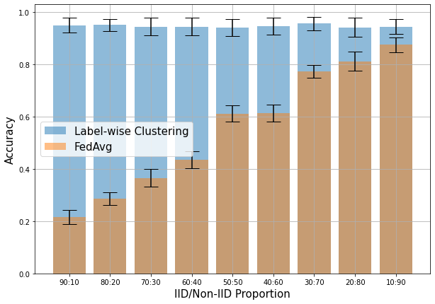}}
\caption{Comparing the accuracy of corresponding proportions of IID : Non-IID in FMNIST dataset.}
\end{figure}

\begin{figure}
\centerline{\includegraphics[width=18.5pc]{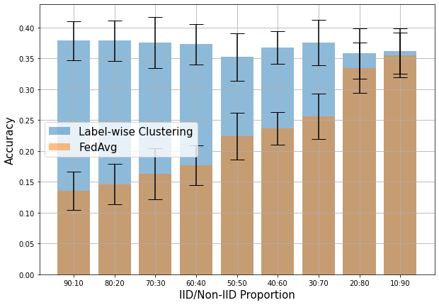}}
\caption{Comparing the accuracy of corresponding proportions of IID : Non-IID in Cifar-10 dataset.}
\end{figure}

\begin{table}[h]
\centering
%\captionsetup{justification=centering}
\caption{TRAIN SUCCESS RATE AMONG PROPORTIONS OF GIVEN IID AND NON-IID IN FEDAVG AND LABEL-WISE CLUSTERING}
\label{table}
\tablefont%
\setlength{\tabcolsep}{3pt}

\begin{tabular}{| c | c | c |}    %\begin{tabular}{c | c | c } = 중간 선 있음
\Xhline{2\arrayrulewidth}
Non-IID & FedAvg SR  & Label-wise Clustering SR \\ [0.8ex] 
\Xhline{2\arrayrulewidth}
90:10 & 0.12  & 1.00 \\ 

80:20  & 0.24 & 1.00 \\

70:30  & 0.34 & 1.00 \\

60:40  & 0.44 & 1.00  \\

50:50  & 0.66 & 1.00 \\ 

60:40  & 0.68 & 1.00  \\ 

30:70  & 0.86 & 1.00\\ 

20:80  & 0.90 & 1.00 \\ 

10:90  & 0.92 & 1.00 \\ [0.75ex] 
\Xhline{2\arrayrulewidth}
\end{tabular}
%\end{center}
\label{tab1}
\end{table}

In addition, we list the train success rate for each proportion $p\left(x_h\right)=\{0.1\cdot h|1\le h\le 9,h\in \mathbb{N}\mathrm{\}}$ in Table II, which we define the success rate as the number of selected clients that contains higher than 20\% of accuracy, out of local models that were selected in 30 trials. Intuitively, success rate approximating to $p\left(x_h\right)$ seems sound, however after 30 trials the overall result shows that it was not directly proportional, and it tends to attain the higher accuracy than given proportion of IID dataset where $\forall SR(x_h)>\forall p\left(x_i\right)$ and $SR(\cdot )$ denotes the success rate. The Pearson correlation coefficient between proportion and FedAvg was 0.98, and the average of MSE $\frac{1}{n}\sum^n_{h=0}{{\left(p\left(x_h\right)-\mathrm{\ }SR\left(x_h\right)\right)}^2}$ of FedAvg was 0.813, which indicates that success rate is not directly proportional to IID ratio among the total local dataset. Values in Table II shows the success rate in given proportions of IID and non-IID in each FedAvg and Label-wise Clustering. 

\footnote{All the codes implemented in this experiment can be found in https://github.com/hml763/Federated-Learning-Label-wise-Clustering.}  

%여기 부터 보기

\section{Conclusion and Future Work}

In this paper, we propose the novel algorithm which discretionally selects and aggregates only the beneficial local models among the FL networks. As the well-known expression garbage in, garbage out illustrates the importance of data quality, the fundamental principle when training an AI model is to fully prepare the cleansed training dataset, which is why most of the data scientists invest their considerable amount of effort on data preprocessing. 

The aim of our algorithm lies within the similar context. We seek breakthroughs of non-convergence in FL with a basic and powerful approach: selecting the local clients that contain various and approximates into uniform distribution of different labels. Due to the diverse environment of each local model, heterogeneity occurs from the non-IID local training dataset, which leads the clients' loss function to go astray as global epoch continues. Furthermore, this triggers the client drift among FL networks, with unstable convergence and substantially degrading the overall performance, as we have defined the six heterogeneous non-IID scenarios and their training experiment results in section III-1. In such cases, our Label-wise clustering algorithm guarantees the loss convergence by agglomerating the locals that were trained with diverse and uniformly distributed labels.

For future works, our interest aims to design the statistical functions of the given non-IID in FL to generalize the models for optimized alternatives, as we have  demonstrated the train feasibility through experiments in section III-1. In this regard, designing the function models contains significant meaning since it could provide vigorous implementation in any FL networks consisting of heterogeneous local clients trained with certain distribution of non-IID dataset. Furthermore, our goal is to discover the explainability of trainability when combining the local models, such as in CNN some models could be trained to classify a cat with its ears, whereas others may emphasize on its eye or other parts. Thus, knowing the impact of aggregating unique models and how to construct the optimal combination is another important agenda to interpret the essence of collaboration in FL. Therefore, our ultimate research value rests on comprehending the overall FL framework to gain the predictability in probable cases and suggest the optimal resolutions.

% \section*{References}

\begin{IEEEbiography}[{\includegraphics[width=1in,height=1.25in,clip,keepaspectratio]{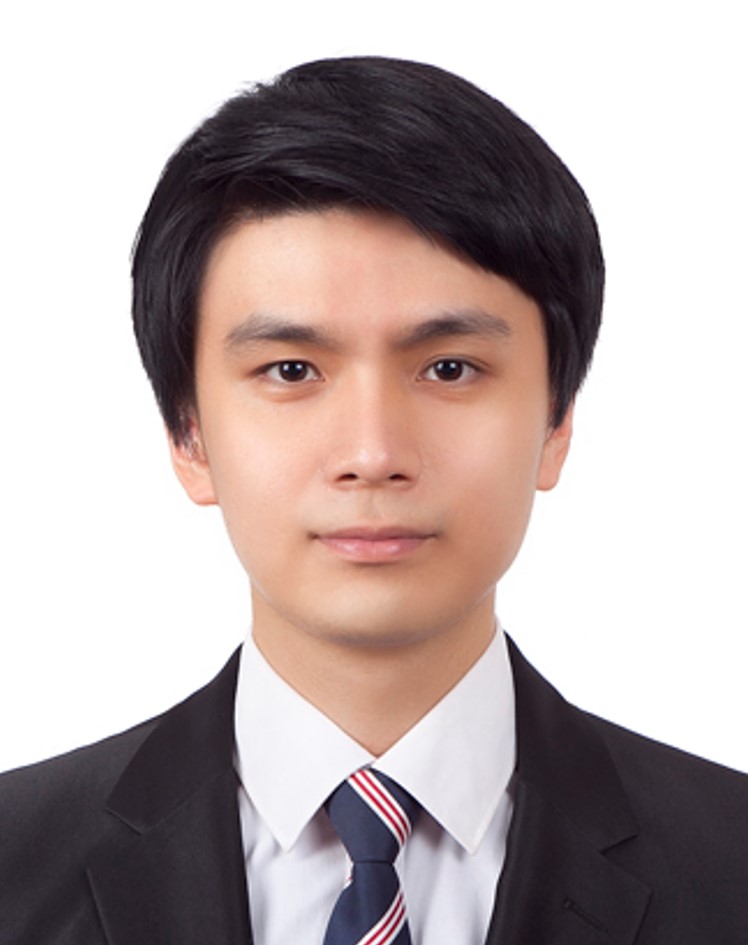}}]{Hunmin Lee}{\space} received the bachelor's degree in computer engineering from Chonnam National University, Gwangju, Rep. of Korea, in 2020. He is currently pursuing his Ph.D. degree in Georgia State University, Atlanta, GA, USA. His current research interests span optimizations in distributed environment (e.g. federated learning), data science, and system design based on AI. Mr. Lee is a 1${}^{st}$ place winner of the 2019 international open data challenge, Tokyo, Japan, and the recipient of the Korean Institute of Smart Media Conference Best Paper Award in 2019.
\end{IEEEbiography}

\begin{IEEEbiography}[{\includegraphics[width=1in,height=1.25in,clip,keepaspectratio]{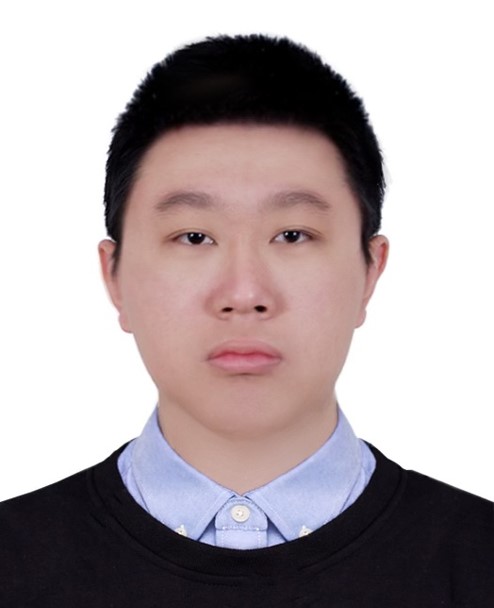}}]{Yueyang Liu}{\space}received his M.S degree in data science from the American University, Washington, D.C., in 2021. He is currently pursuing the Ph.D. degree in computer science at Georgia State University. His research interests include data mining, NLP, data security, abnormal detection  and machine learning.
\end{IEEEbiography}

\begin{IEEEbiography}[{\includegraphics[width=1in,height=1.25in,clip,keepaspectratio]{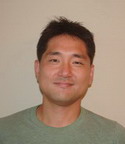}}]{Donghyun Kim}{\space}(Senior Member, IEEE)
received the B.S. degree in electronic and computer engineering, and the M.S. degree in computer science and engineering from Hanyang University, Ansan, South Korea, in February 2003 and February 2005, respectively, and the Ph.D. degree in computer science from the University of Texas at Dallas, Richardson, TX, USA, in May 2010. He is currently an Assistant Professor with the Department of Computer Science, Georgia State University (GSU), Atlanta, GA, USA. He is a Senior Member of ACM. He has served as a TPC co-chair for several international conferences, most recently IPCCC 2020 and COCOON 2020.
\end{IEEEbiography}

\begin{IEEEbiography}[{\includegraphics[width=1in,height=1.25in,clip,keepaspectratio]{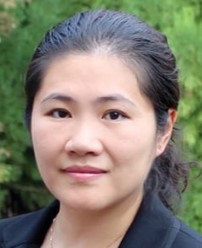}}]{Yingshu Li}{\space} received her Ph.D. and M.S. degrees from the Department of Computer Science and Engineering at University of Minnesota-Twin Cities. Dr. Li is currently a Professor in the Department of Computer Science and an affiliated faculty member in the INSPIRE Center at Georgia State University. Her research interests include Privacy-aware Computing, Management of Big Sensory Data, Internet of Things, Social Networks, and Wireless Networking. Dr. Li is the recipient of the NSF CAREER Award.
\end{IEEEbiography}

\end{document}